\documentclass[10pt,twocolumn,letterpaper]{article}

\usepackage{cite}
\usepackage{cvpr}
\usepackage{times}
\usepackage{epsfig}
\usepackage{graphicx}
\usepackage{amsmath}
\usepackage{amssymb}
\usepackage{booktabs}
\usepackage{caption}

\usepackage[pagebackref=true,breaklinks=true,letterpaper=true,colorlinks,bookmarks=false]{hyperref}

\cvprfinalcopy 

\def\cvprPaperID{1926} 

\DeclareMathOperator{\E}{\mathbb{E}}
\makeatletter
\newcommand\footnoteref[1]{\protected@xdef\@thefnmark{\ref{#1}}\@footnotemark}
\makeatother
\newcommand{\new}[1]{{\color[rgb]{0,0,1} #1}}

\ifcvprfinal\fi
\begin{document}


\title{Actor and Observer: Joint Modeling of First and Third-Person Videos}

\newcommand{\myurl}[1]{\url{#1}}
\renewcommand{\thefootnote}{\fnsymbol{footnote}}
\author{Gunnar A. Sigurdsson $^1\footnotemark[1]$ \ \ \ \ 
Abhinav Gupta $^{1}$ \ \ \ 
Cordelia Schmid $^{2}$ \ \ \ 
Ali Farhadi $^{3}$ \ \ \ 
Karteek Alahari $^{2}$ \\
$^1$Carnegie Mellon University~~~~ \  
$^2$Inria$\footnotemark[2]$~~~~ \ \ 
$^3$Allen Institute for Artificial Intelligence \\
\myurl{github.com/gsig/actor-observer}\vspace{-0.5cm}
}


\thispagestyle{empty}

\twocolumn[{%
\renewcommand\twocolumn[1][]{#1}%
\maketitle
\thispagestyle{empty}
\begin{center}
    \centering
    \includegraphics[width=1.0\linewidth]{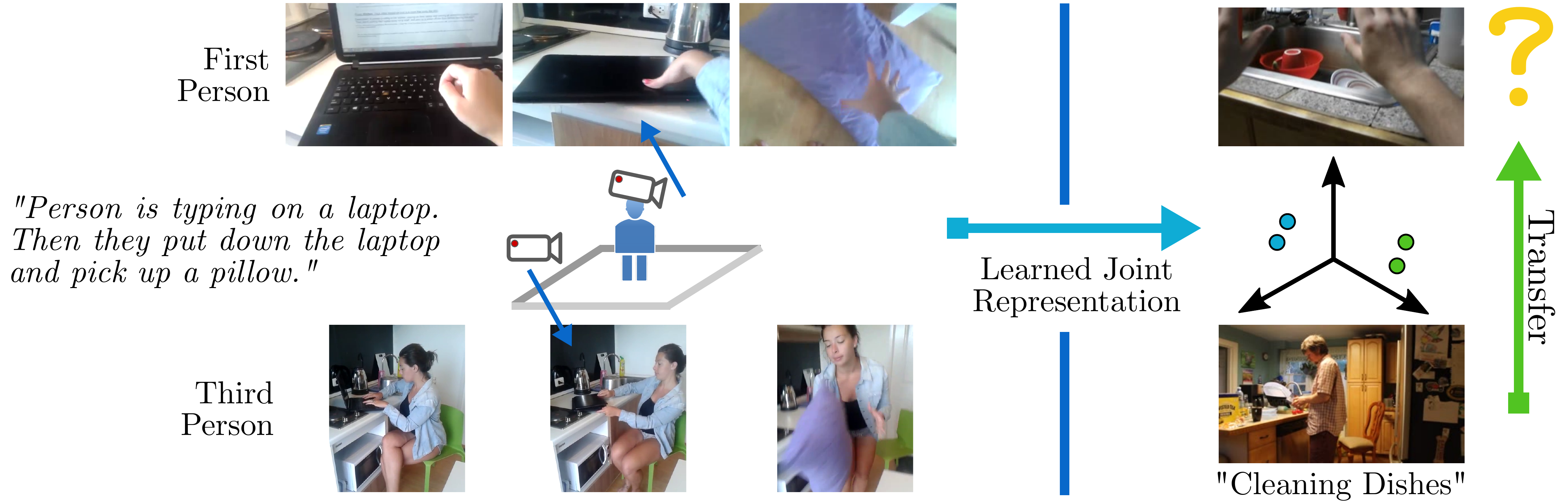}
    \vspace{-1.5em}
    \captionof{figure}{We explore how to reason jointly about first and third-person for understanding human actions. We collect paired data of first and third-person actions sharing the same script. Our model learns a representation from the relationship between these two modalities. We demonstrate multiple applications of this research direction, for example, transferring knowledge from the observer's to the actor's perspective.\label{fig:teaser}}
\end{center}%
}]

\begin{abstract}
\vspace{-1em}
Several theories in cognitive neuroscience suggest that when people interact with the world, or simulate interactions, they do so from a first-person egocentric perspective, and seamlessly transfer knowledge between third-person (observer) and first-person (actor). Despite this, learning such models for human action recognition has not been achievable due to the lack of data. This paper takes a step in this direction, with the introduction of Charades-Ego, a large-scale dataset of paired first-person and third-person videos, involving 112 people, with 4000 paired videos. This enables learning the link between the two, actor and observer perspectives. Thereby, we address one of the biggest bottlenecks facing egocentric vision research, providing a link from first-person to the abundant third-person data on the web. We use this data to learn a joint representation of first and third-person videos, with only weak supervision, and show its effectiveness for transferring knowledge from the third-person to the first-person domain.
\end{abstract}
\vspace{-3em}

\footnotetext{\footnotemark[1]Work was done while Gunnar was at Inria.}
\footnotetext{\footnotemark[2]Univ.\ Grenoble Alpes, Inria, CNRS, Grenoble INP, LJK, 38000 Grenoble, France.}

\section{Introduction}
\vspace{-0.15cm}
What is an action? How do we represent and recognize actions? Most of the current research has focused on a data-driven approach using abundantly available third-person (observer's perspective) videos. But can we really learn how to represent an action without understanding goals and intentions? Can we learn goals and intentions without simulating actions in our own mind? A popular theory in cognitive psychology, the Theory of Mind~\cite{premack_woodruff_1978}, suggests that humans have the ability to put themselves in each others' shoes, and this is a fundamental attribute of human intelligence. In cognitive neuroscience, the presence of activations in mirror neurons and motor regions even for passive observations suggests the same\cite{rizzolatti2004mirror}.

When people interact with the world (or simulate these interactions), they do so from a first-person egocentric perspective~\cite{kanade2012first}. Therefore, making strides towards human-like activity understanding might require creating a link between the two worlds of data: first-person and third-person. In recent years, the field of egocentric action understanding~\cite{lee2012discovering,Li_2015_CVPR,pirsiavash2012detecting,ryoo2013first,jayaraman2015learning,Rhinehart_2017_ICCV} has bloomed due to a variety of practical applications, such as augmented/virtual reality. While first-person and third-person data represent the two sides of the same coin, these two worlds are hardly connected. Apart from philosophical reasons, there are practical reasons for establishing this connection. If we can create a link, then we can use billions of easily available third-person videos to improve egocentric video understanding. Yet, there is no connection: why is that?

The reason for the lack of link is the lack of data! In order to establish the link between the first and third-person worlds, we need aligned first and third-person videos. In addition to this, we need a rich and diverse set of actors and actions in these aligned videos to generalize. As it turns out, aligned data is much harder to get. In fact, in the egocentric world, getting diverse actors and, thus, a diverse action dataset is itself a challenge that has not yet been solved. Most datasets are lab-collected and lack diversity as they contain only a few subjects~\cite{fathi2011learning,pirsiavash2012detecting,firstthird2017cvpr}.

\begin{figure*}[t]
    \centering
    \includegraphics[width=1.0\linewidth]{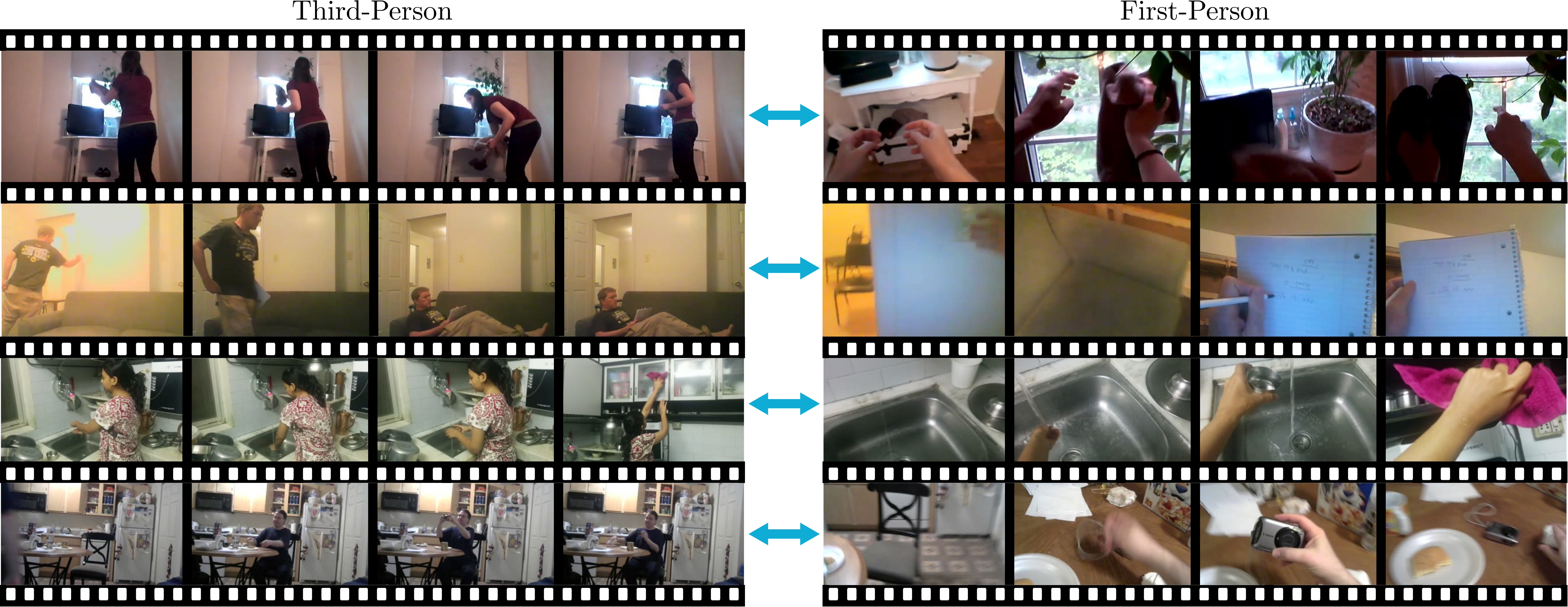}
    \vspace{-0.7cm}
    \caption{Examples from Charades-Ego, showing third-person (left) and the corresponding first-person (right) video frames.\vspace{-0.5cm}}
    \label{fig:datateaser}
\end{figure*}

In this paper, we address one of the biggest bottlenecks facing egocentric vision research. We introduce a large-scale and diverse egocentric dataset, Charades-Ego, collected using the Hollywood in Homes~\cite{charades} methodology. We demonstrate an overview of the data collection and the learning process in Figure~\ref{fig:teaser}, and present examples from the dataset in Figure~\ref{fig:datateaser}. Our new dataset has 112 actors performing 157 different types of actions. More importantly, we have the same actors perform the same sequence of actions from both first and third-person perspective. Thus, our dataset has semantically similar first and third-person videos. These ``aligned'' videos allow us to take the first steps in jointly modeling actions from first and third-person's perspective. Specifically, our model, ActorObserverNet, aligns the two domains by learning a joint embedding in a weakly-supervised setting. We show a practical application of joint modeling: transferring knowledge from the third-person domain to the first-person domain for the task of zero-shot egocentric action recognition.

\subsection{Related work} 
\vspace{-0.15cm}
Action recognition from third-person perspective has been extensively studied in computer
vision. The most common thread is to use hand-crafted features~\cite{STIP05,HOG3D,HOF} or learn features for recognition using large-scale datasets~\cite{i3D,simonyan2014twostream}. We refer the reader to~\cite{poppe2010survey,weinland2011survey} for a detailed survey of these approaches, and in the following we focus on the work most relevant to our approach. Our work is inspired by methods that attempt to go beyond modeling appearances~\cite{wang2015unsupervised,jayaraman2015learning}. Our core hypothesis is that modeling goals and intentions requires looking beyond the third-person perspective.\\
{\bf Egocentric understanding of activities.}
Given recent availability of head-mounted cameras of various types, there has been a significant amount of work in understanding first-person egocentric data~\cite{fathi2011understanding,lee2012discovering,Li_2015_CVPR,pirsiavash2012detecting,ryoo2013first,ma2016going}. This unique insight into people's behaviour gives rise to interesting applications such as predicting where people will look~\cite{Li_2015_CVPR}, and how they will interact with the environment~\cite{rhinehart2016learning}.
Furthermore, it has recently been shown that egocentric training data provides strong features for tasks such as object detection~\cite{jayaraman2015learning}.\\
{\bf Datasets for egocentric understanding.}
Egocentric video understanding has unique challenges as
datasets~\cite{fathi2011learning,pirsiavash2012detecting,lee2012discovering,firstthird2017cvpr} are smaller by an order of magnitude than their third-person equivalents~\cite{caba2015activitynet,charades}. This is due to numerous difficulties in collecting such data, e.g., availability, complexity, and privacy. Recent datasets have targeted this issue by using micro-videos from the internet, which include both third and first-person videos~\cite{nguyen2016open}. While they contain both first and third-person videos, there are no paired videos that can be used to learn the connection between these two domains. In contrast, our dataset contains corresponding first and third-person data, enabling a joint study.\\
{\bf Unsupervised and self-supervised representation learning.}
In this work, we use the multi-modal nature of the data to learn a robust representation across those modalities. It allows us to learn a representation from the data alone, without any explicit supervision. This draws inspiration from recent work on using other cues for representation learning, such as visual invariance for self-supervised learning of features~\cite{wang2015unsupervised,wang2015deep,Srivastava2015,jayaraman2015learning,Agrawal2015,Mathieu2015,Li2016,Pathak2017}. For example, this visual invariance can be obtained by tracking how objects change in videos~\cite{wang2015unsupervised} or from consecutive video frames~\cite{Mathieu2015}.
Typically, this kind of invariance is harnessed via deep metric learning with Siamese (triplet) architectures~\cite{Chopra2005,Hadsell2006,Gong2013,Wang2014,zagoruyko2015learning,hoffer2016deep}.\\
{\bf Data for joint modeling of first and third person.}
To learn to seamlessly transfer between the first and third-person perspectives we require paired data of these two domains. Some recent work has explored data collected from multiple viewpoints for a fine-grained understanding human actions~\cite{joo_iccv_2015}. Due to the difficulty of acquiring such data, this is generally done in a small-scale lab setting~\cite{firstthird2017cvpr,joo_iccv_2015}, with reconstruction using structure-from-motion techniques~\cite{joo_iccv_2015}, or matching camera and head motion of the exact same event~\cite{poleg2014head,yonetani2015ego}. Most related to our work is that of Fan et al.~\cite{firstthird2017cvpr} which collects 7 pairs of videos in a lab setting, and learns to match camera wearers between third and first-person. In contrast, we look at thousands of diverse videos collected by people in their homes.

\vspace{-0.2cm}
\section{Charades-Ego}
\vspace{-0.2cm}
In order to link first-person and third-person data, we need to build a dataset that has videos shot in first and third-person views. We also need the videos to be semantically aligned, i.e., the same set of actions should appear in each video pair. Collection in a controlled lab setting is difficult to scale, and very few pairs of videos of this type are available on the web. In fact, collection of diverse egocentric data is a big issue due to privacy concerns. So how do we scale such a collection?

We introduce the Charades-Ego dataset in this paper. The dataset is collected by following the methodology outlined by the ``Hollywood in Homes'' approach~\cite{charades}, originally used to collect the Charades dataset~\cite{charades,challenge}, where workers on Amazon Mechanical Turk (AMT) are incentivized to record and upload their own videos. This in theory allows for the creation of any desired data.

In particular, to get data that is both in first and third-person we use publicly available scripts from the Charades dataset~\cite{charades}, and ask users to record two videos: (1) one with them acting out the script from the third-person; and (2) another one with them acting out the same script in the same way, with a camera fixed to their forehead. We ensure that all the 157 activity classes from Charades occur sufficiently often in our data. The users are given the choice to hold the camera to their foreheads, and do the activities with one hand, or create their own head mount and use two hands. We encouraged the latter option by incentivizing the users with an additional bonus for doing so.\footnote{We compensated AMT workers \$1.5 for each video pair, and \$0.5 in additional bonus.} This strategy worked well, with 59.4\% of the submitted videos containing activities featuring both hands, courtesy of a home-made head mount holding the camera. 

Specifically, we collected 4000\footnote{Since the scripts are from the Charades dataset, each video pair has another third-person video from a different actor. We use this video also in our work.} pairs of third and first-person videos (8000 videos in total), with over 364 pairs involving more than one person in the video. The videos are $31.2$ seconds long on average. This data contains videos that follow the same structure semantically, i.e., instead of being identical, each video pair depicts activities performed by the same actor(s) in the same environment, and with the same style. This forces a model to latch onto the semantics of the scene, and not only landmarks. We evaluated the alignment of videos by asking workers to identify moments that are shared across the two videos, similar to the algorithmic task in Section~\ref{sec:align}, and found the median alignment error to be 1.3s (2.1s average). This offers a compromise between a synchronized lab setting to record both views simultaneously, and scalability. In fact, our dataset is one of the largest first-person datasets available~\cite{fathi2011learning,lee2012discovering,pirsiavash2012detecting,firstthird2017cvpr}, has significantly more diversity (112 actors in many rooms), and most importantly, is the only large-scale dataset to offer pairs of first and third-person views that we can learn from. Examples from the dataset are presented in Figure~\ref{fig:datateaser}. Our data is publicly available at \myurl{github.com/gsig/actor-observer}. 

\vspace{-0.2cm}
\section{Jointly Modeling First and Third-Person}
\vspace{-0.2cm}
As shown in Figure~\ref{fig:teaser}, our aim is to learn a shared representation, i.e., a common embedding for data, from the corresponding frames of the first and the third-person domains. In the example in the figure, we have a full view of a person working on a laptop in third-person. We want to learn a representation where the corresponding first-person view, with a close-up of the laptop screen and a hand typing, has a similar representation. We can use the correspondence between first and third-person as supervision to learn this representation that can be effective for multiple tasks. The challenges in achieving this are: the views are very visually different, and many frames are uninformative, such as walls, doors, empty frames, and blurry frames. We now describe a model that tackles these challenges by learning how to select training data for learning a joint representation.

\vspace{-0.1cm}
\subsection{Formulation}
\label{sec:formulation}
\vspace{-0.1cm}
The problem of modeling the two domains is a multi-modal learning problem, in that, data in the first-person view is distinct from data in the third-person view. Following the taxonomy of Baltrusaitis et al.~\cite{baltruvsaitis2017multimodal} we formulate this as learning a \emph{coordinated representation} such that corresponding samples in both the first and third-person modalities are close-by in the joint representation. The next question is how to find the alignment or corresponding frames between the two domains. We define ground-truth alignment as frames from first and third-person being within $\Delta$-seconds of each other, and non-alignment as frames being further than $\Delta'$-seconds, to allow for a margin of error.

If a third-person frame $x$ and a first-person frame $z$ map to representations $f(x)$ and $g(z)$ respectively, we want to encourage similarity between $f(x){\sim}g(z)$ if their timestamps $t_x$ and $t_z$ satisfy $|t_x-t_z|<\Delta$. If the two frames do not correspond, then we maximize the distance between their learned representations $f(x)$ and $g(z)$. One possible way to now learn a joint representation is to sample all the corresponding pairs of $(x,z)$, along with a non-corresponding first-person frame $z'$ and use a triplet loss. However, this is not ideal for three reasons: (1)~It is inefficient to sample all triplets of frames; (2)~Our ground truth (correspondence criteria) is weak as videos are not perfectly synchronized. (3)~We need to introduce a mechanism which selects samples that are informative (e.g., hand touching the laptop in Figure~\ref{fig:teaser}) and conclusive. These informative samples can also be non-corresponding pairs (negative).

We define the problem of learning the joint representation formally with our loss function $l_{\theta}$. The loss is defined over triplets from the two modalities $(x{,}z{,}z')$. The overall objective function is given by:
\begin{equation}
    \vspace{-0.1cm}
    L = \mathop{\E}_{(x{,}z{,}z') \sim P_{\theta}} \left[ l_{\theta}(x{,}z{,}z') \right],
    \vspace{-0.1cm}
    \label{eq:originalloss}
\end{equation}
where $l_\theta$ is a triplet loss on top of ConvNet outputs, and $\theta$ is set of all the model parameters. The loss is computed over a \emph{selector} $P_\theta$. We also learn $P_\theta$, a parameterized discrete distribution over data, that represents how to sample more informative data triplets $(x{,}z{,}z')$. Intuitively, this helps us find what samples are likely too hard to learn from. To avoid the degenerate solution where $P_\theta$ emphasizes only one sample, we constrain $P_\theta$ by reducing the complexity of the function approximator, as discussed in Section~\ref{sec:optimize}.

\begin{figure}[tb]
    \centering
    \includegraphics[width=1.0\linewidth]{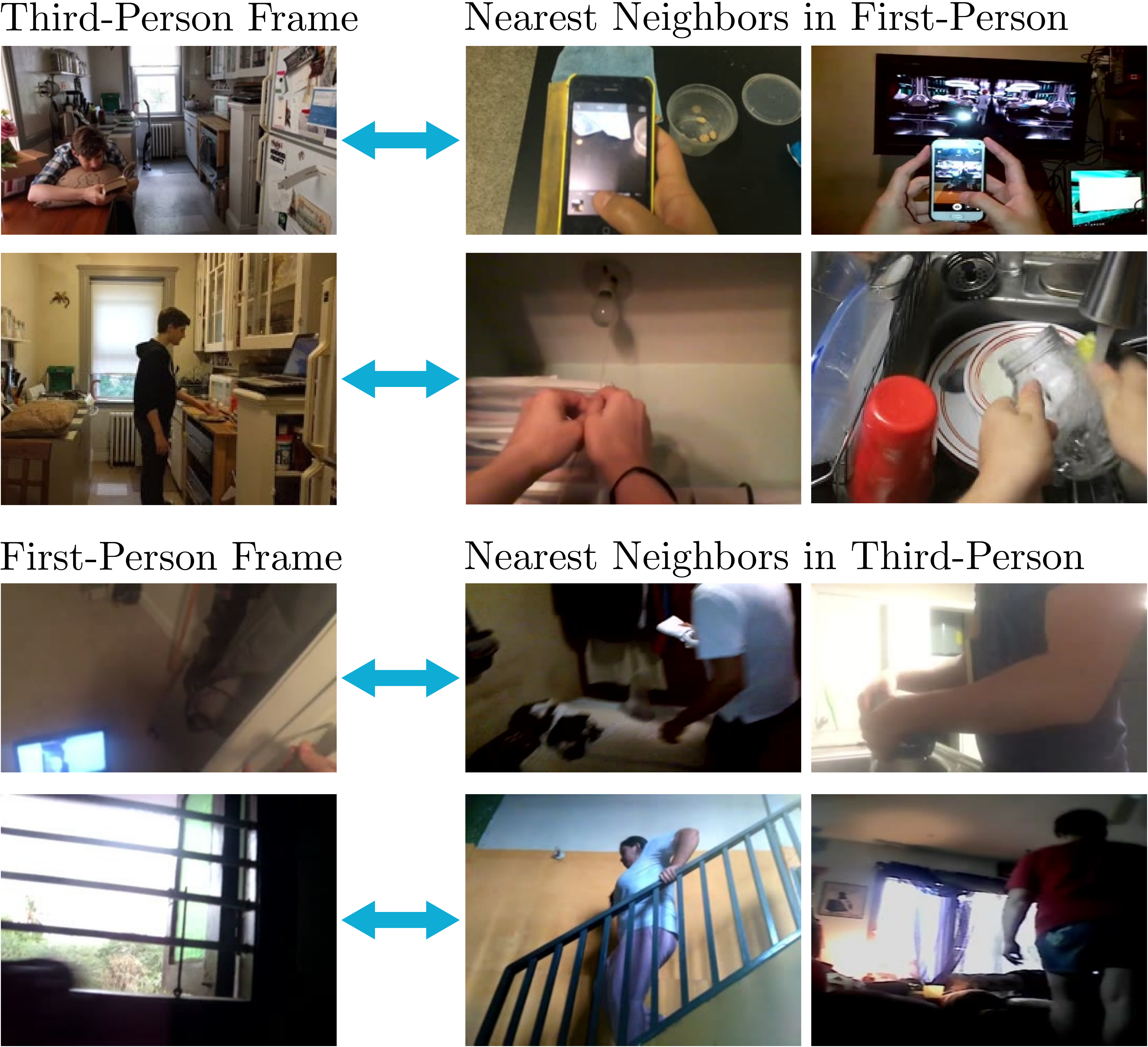}
    \vspace{-0.6cm}
    \caption{Using our joint first and third-person model we can hallucinate how a scene might look through the eyes of the actor in the scene. The top two rows show nearest neighbours (on the right) from first-person videos. The bottom two rows show the observer's perspective, given a first-person video frame.\vspace{-0.3cm}}
    \label{fig:nn}
\end{figure}

The joint model from optimizing the loss and the selector can be used to generate the other view, given either first or third-person view. We illustrate this in Figure~\ref{fig:nn}, where we find the closest first-person frames in the training set, given a third-person query frame. We see that the model is able to connect the two views from the two individual frames, and hallucinate what the person is seeing. 

Our setup is related to previous formulations in self-supervised and unsupervised learning, where the pairs $(x{,}z)$ are often chosen with domain-specific heuristics, e.g., temporal~\cite{jayaraman2015learning,wang2015unsupervised} and spatial~\cite{doersch2015unsupervised} proximity. Triplet loss is a common choice for the loss $l_\theta$ for these tasks~\cite{jayaraman2015learning,wang2015unsupervised,doersch2015unsupervised,hoffer2016deep}. We will now address how we model our loss function with a ConvNet, and optimize it with stochastic gradient descent.

\subsection{Optimizing the objective}
\label{sec:optimize}
\vspace{-0.1cm}
Optimizing the objective involves learning parameters of both the triplet loss $l_\theta$, as well as the selector $P_\theta$. This correlated training can diverge. We address this by using importance sampling to rewrite the objective $L$ (\ref{eq:originalloss}) to an equivalent form. We move the distribution of interest $P_\theta$ to the objective and sample from a different fixed distribution $Q$ as follows: 
\begin{equation} 
    L = \mathop{\E}_{(x{,}z,{z'}) \sim Q} \left[ \frac{p_{\theta}(x{,}z{,}z')}{q(x{,}z{,}z')} l_{\theta}(x{,}z{,}z') \right]. \label{eq:finalloss}
\end{equation}
We choose $Q$ to be a uniform distribution over the domain of possible triplets: $\{(x,z,z') \mid |t_x{-}t_z|{<}\Delta, |t_x{-}t_z'|{>}\Delta'\}$. We uniformly sample frames from first and third-person videos, but re-weight the loss based on the informativeness of the triplet. Here, $p_\theta(x,z,z')$ is the value of the selector for the triplet choice $(x,z,z')$.

Instead of modeling the informativeness of the \emph{whole} triplet, we make a simplifying assumption. We assume the selector $P_\theta$ factorizes as $p_\theta(x{,}z{,}z'){=}p_\theta(x)p_\theta(z)p_\theta(z')$. Further, we constrain $P_\theta$ such the probability of selecting any given frame in that video sums to one for a given video. This has similarities with the concept of ``bags'' in multiple instance learning~\cite{andrews2003support}, where we only know whether a given set (bag) of examples contains positive examples, but not if all the examples in the set are positive. Similarly, here we learn a distribution that determines how to select the useful examples from a set, where our sets are videos. We use a ConvNet architecture to realize our objective.

\subsection{Architecture of ActorObserverNet}
\begin{figure}[t]
    \centering
    \includegraphics[width=1.0\linewidth]{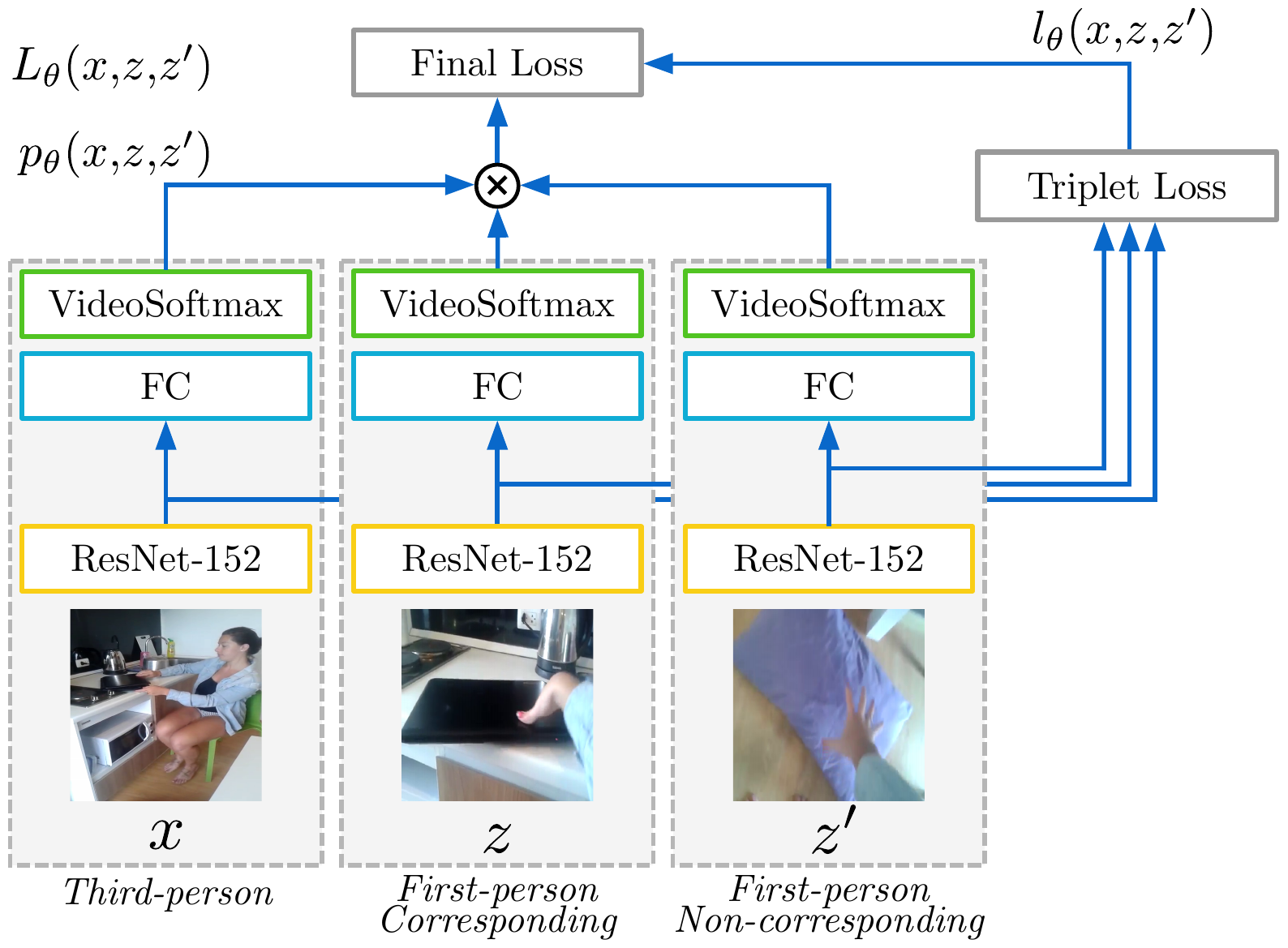}
    \caption{Illustration of our ActorObserverNet. The model has separate streams for first and third-person. Given a triplet of frames from these two modalities, the model computes their fc7 features, which are used to compare and learn their similarity. The FC and the VideoSoftmax layers also compute the likelihood of this sample with respect to the selector $P_\theta$.\vspace{-0.5cm}}
    \label{fig:model}
\end{figure}

\vspace{-0.1cm}
The ConvNet implementation of our model is presented in Figure~\ref{fig:model}. It consists of three streams: one for third-person, and two for first-person with some shared parameters. The streams are combined with a L2-based distance metric~\cite{hoffer2016deep} that enforces small distance between corresponding samples, and large distance between non-corresponding ones:
\newcommand{\norm}[1]{\left\lVert#1\right\rVert}
\begin{align}
\vspace{-0.1cm}
    l_\theta(x{,}z{,}z') &= \frac{e^{\norm{x-z}_2}}{e^{\norm{x-z}_2}+e^{\norm{x-z'}_2}}.
\vspace{-0.1cm}
\end{align}

The computation of the selector value, $p_\theta(x{,}z{,}z')$, for a triplet $(x{,}z{,}z')$ is also done by the three streams. The selector values are the result of a $4096{\times}1$ fully-connected layer, followed by a scaled tanh nonlinearity\footnote{The choice of Tanh nonlinearity makes the network more stable than unbounded alternatives like ReLU.} for each stream. We then define a novel non-linearity, VideoSoftmax, to compute the per-video normalized distribution over frames in different batches, which are then multiplied together to form $p_\theta(x)p_\theta(z)p_\theta(z')$. Once we have the different components of the loss in (\ref{eq:finalloss}) we add a loss layer (``Final loss'' in the figure). This layer combines the triplet loss $l_\theta$ with the selector output $p_\theta$ and implements the loss in (\ref{eq:finalloss}). All the layers are implemented to be compatible with SGD~\cite{asynchronous}. More details are provided in the supplementary material.

\vspace{0.1cm}
\noindent{\bf VideoSoftmax layer.} The distribution $P_\theta$ is modeled with a novel layer which computes a probability distribution across multiple samples corresponding to the same video, even if they occur in different batches. The selector value for a frame $x$ is given by:
\begin{equation}
\vspace{-0.1cm}
    p_\theta(x) = \frac{e^{f_\theta(x)}}{\sum\limits_{x' \in \mathcal{V}} e^{f_\theta(x')}},
\vspace{-0.1cm}
\label{eq:softmax}
\end{equation}
where $f_\theta(x)$ is the input to the layer and denominator is the sum of $e^{f_\theta(x')}$ computed over all frames $x'$ in the same video $\mathcal{V}$. This intuitively works like a softmax function, but across frames in the same video.

Since triplet loss $l_\theta$ is weighted by the output of the selector, the gradient updates with respect to the triplet loss are simply a weighted version of the original gradient. The gradient for optimizing the loss in (\ref{eq:finalloss}) with respect to the selector in (\ref{eq:softmax}) is (with slight abuse of notation for simplicity):
\begin{equation}
    \frac{\partial L}{\partial f} \propto p_{\theta}(x{,}z{,}z')(l_{\theta}(x{,}z{,}z')-L),
\end{equation}
where the gradient is with respect to the input of the VideoSoftmax layer $f$, so we can account for the other samples in the denominator of (\ref{eq:softmax}). $Q$ is defined as a constant over the domain, and can be ignored in the derivation. The intuition is that this decreases the weight of the samples that are above the loss $L$ (\ref{eq:originalloss}), and increases it otherwise. Our method is related to mining easy examples. The selector learns to predict the relative weight of each triplet, i.e., instead of using the loss directly to select triplets (as in mining hard examples). The gradient is then scaled by the magnitude of the weight. The average loss $L$ is computed across all the frames; see
supplementary material for more details.

\vspace{-0.2cm}
\section{Experiments}
\label{sec:expts}
\vspace{-0.2cm}
We demonstrate the effectiveness of our joint modeling of first and third-person data through several applications, and also analyze what the model is learning. In Section~\ref{sec:map3to1} we evaluate the ability of the joint model to discriminate correct first and third-person pairs from the incorrect ones. We investigate how well the model localizes a given first-person moment in a third-person video, from the same as well as users, by temporally aligning a one-second moment between the two videos (Section~\ref{sec:align}). Finally, in Section~\ref{sec:zeroshot} we present results for transferring third-person knowledge into the first-person modality, by evaluating zero-shot first-person action recognition. We split the 8000 videos into $80\%$ train/validation, and $20\%$ test for our experiments.

\begin{figure*}[tb]
    \centering
    \includegraphics[width=1.0\linewidth]{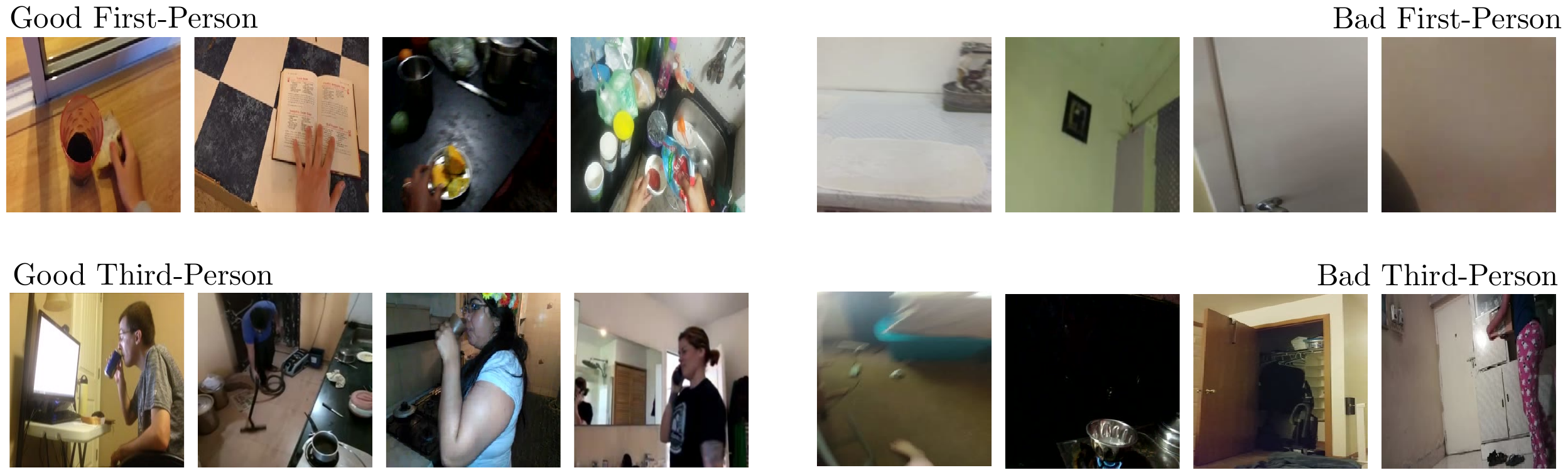}
    \vspace{-0.5cm}
    \caption{A selection of frames, from third and first-person videos, the model assigns the highest and the lowest weights, i.e., $p_\theta(x)$ and $p_\theta(z)$ from (\ref{eq:finalloss}) respectively. This provides intuition into what the model is confident to learn from.\vspace{-0.2cm}}
    \label{fig:goodbad}
\end{figure*}
\subsection{Implementation details}
Our model uses a ResNet-152 
video frame classification architecture, pretrained
on the Charades dataset~\cite{charades}, and shares parameters between both the first and third-person streams. This is inspired by the two-stream model~\cite{simonyan2014twostream}, which is a common baseline architecture even in ego-centric videos~\cite{ma2016going,firstthird2017cvpr}. The scale of random crops for data augmentation in training was set to $80{-}100\%$ for first-person frames, compared to the default $8{-}100\%$ for third-person frames. We set the parameter $\Delta$ for the maximum distance to determine a positive pair as one second (average alignment error in the dataset), and the parameter $\Delta'$ for the negative pair as 10 seconds. More details about the triplet network are available in the supplementary material.

We sample the training data triplets, in the form of a positive pair with  first and third-person frames, which correspond to each other, and a negative pair with the same third-person frame and an unrelated first-person frame from the same video. This sampling is done randomly following the uniform distribution $Q$ in (\ref{eq:finalloss}). The scales of tanh are constrained to be positive. For the experiments in Sections~\ref{sec:align} and~\ref{sec:zeroshot}, the parameters of the fully connected layers for the two first-person streams are shared. Our code is implemented in the PyTorch machine learning framework and is available at \myurl{github.com/gsig/actor-observer}.

\subsection{Mapping third-person to first-person}
\label{sec:map3to1}
The first problem we analyze is learning to model first and third-person data jointly, which is our underlying core problem. We evaluate the joint model by finding a corresponding first-person frame, given a third-person frame, under two settings: (1)~using the whole test set (`All test data'); and (2)~when the model assigns weights to each sample (`Choose $X\%$ of test data'). In the second case, the triplets with the top $5\%$, $10\%$, or $50\%$ highest weights are evaluated. Each triplet contains a given third-person frame, and a positive and negative first-person frames. This allows the model to choose examples from the test set to evaluate.

\new{
\begin{table}[tb]
\begin{center}
\resizebox{\linewidth}{!}{%
\begin{tabular}{@{}lcccc@{}}
\toprule
 & Random & \begin{tabular}[c]{@{}l@{}}ImageNet\\ ResNet-152\end{tabular} & \begin{tabular}[c]{@{}l@{}}Charades\\ Two-Stream\end{tabular} & ActorObserverNet \\ \midrule
\textbf{Same person} & & & & \\
All test data                    & 50.0 & 53.6    & 55.5    & 51.7 \\
Choose 50\%  of test data        & 50.0 & 55.7    & 60.2    & \bf{73.9} \\ 
Choose 10\%  of test data        & 50.0 & 57.9    & 68.8    & \bf{97.2}  \\ 
Choose \enspace 5\% of test data & 50.0 & 56.5    & 71.9    & \bf{96.8} \vspace{0.3em} \\ 
\textbf{Different persons} & & & & \\
All test data                    & 50.0 & 50.6 & 51.7 & 50.4 \\
Choose 50\%  of test data        & 50.0 & 50.4 & 51.6 & \bf{76.3} \\ 
Choose 10\%  of test data        & 50.0 & 49.6 & 50.8 & \bf{98.8}  \\ 
Choose \enspace 5\% of test data & 50.0 & 45.6 & 51.4 & \bf{98.3} \\ \bottomrule

\end{tabular}%
}
\end{center}
\vspace{-0.5cm}
\caption{Given a third-person frame, we determine whether a first-person frame corresponds to it. Results are shown as correspondence classification accuracy (in \%). Higher is better. See Section~\ref{sec:map3to1} for details.}
\label{tbl:firstthird}
\end{table}
}
From Table~\ref{tbl:firstthird} we see that the original problem (`All test data') is extremely challenging, even for state-of-the-art representations. The baseline results are obtained with models using fc7 features from either ResNet-152 trained on ImageNet or a two-stream network (RGB stream using ResNet-152 from~\cite{charades}) trained on Charades to compute the loss. The baselines use the difference in distance between positive and negative pairs as the weight used to pick what samples to evaluate on in the second setting.

The results of the two-Stream network (`Charades Two-Stream' in the table) and our ActorObserverNet using all test data (`All test data') are similar, but still only slightly better than random chance. This is expected, since many of the frames correspond to occluded human actions, people looking at walls, blurry frames, etc., as seen in Figure~\ref{fig:goodbad}. On the other hand, our full model, which learns to weight the frames (`Choose $X\%$ of test data' in the table), outperforms all the other methods significantly. Note that our model assigns a weight for each image frame independently, and in essence, learns if it is a good candidate for mapping. We observe similar behavior when we do the mapping with third and first-person videos containing the same action performed by different people (`Different persons' in the table).

\begin{figure}[tb]
    \centering
    \includegraphics[width=1.0\linewidth]{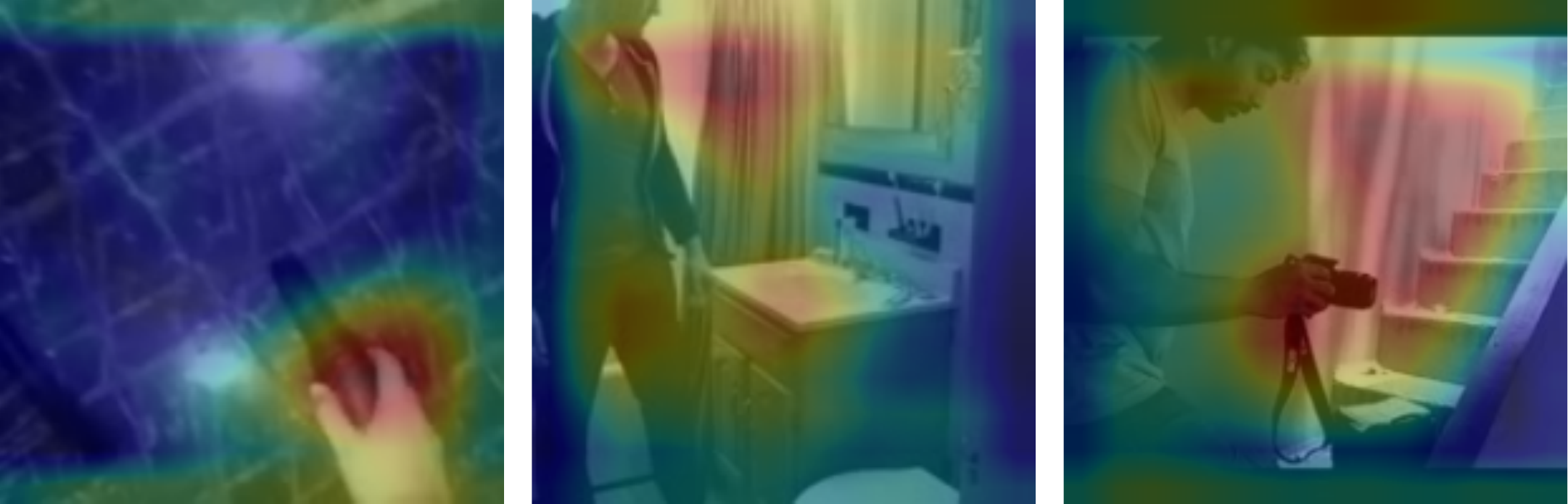}
    \caption{Conv5 activations of ActorObserverNet. The colors range from blue to red, denoting low to high activations. We observe the network attending to hands, objects, and the field of view.}
    \label{fig:attention}
\end{figure}
\begin{figure}[tb]
    \centering
    \includegraphics[width=1.0\linewidth]{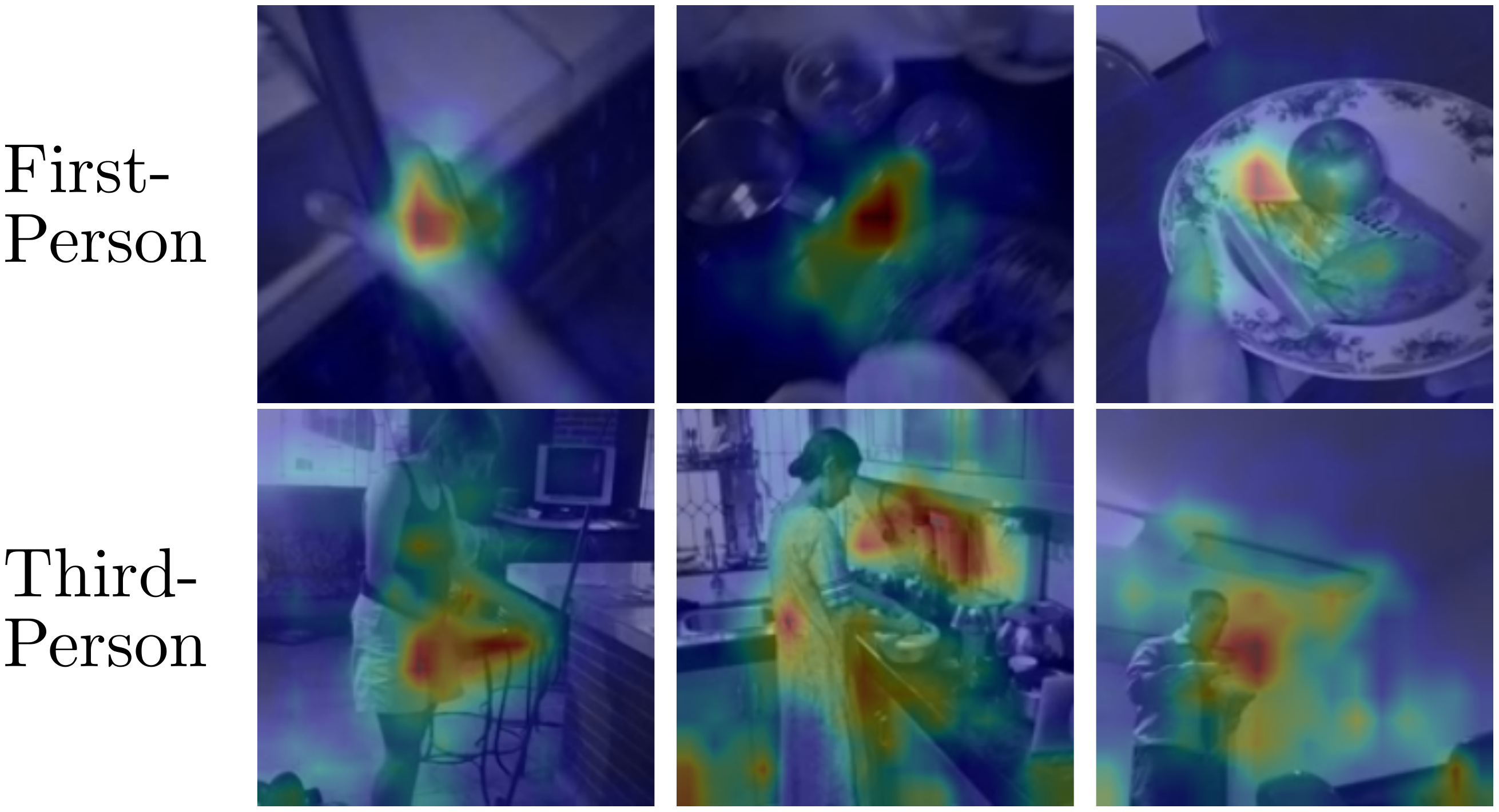}
    \vspace{-0.5cm}
    \caption{By backpropagating the similarity loss to the image layer, we can visualize what regions the model is learning from. The colors range from blue to red, denoting low to high importance.\vspace{-0.3cm}}
    \label{fig:attentionpairs}
\end{figure}
Figure~\ref{fig:goodbad} shows a qualitative analysis to understand what the model is learning. Here, we illustrate the good and the bad frames chosen by the model, according to the learned weights, both in the third and first-person cases. We observe that the model learns to ignore frames without objects and people, and blurry, feature-less frames, such as the ones seen in the bottom row in the figure. Furthermore, our model prefers first-person frames that include hands, and third-person frames with the person performing an action, such as answering a phone or drinking; see frames in the top row in the figure.

\begin{figure}[tb]
    \centering
    \includegraphics[width=1.0\linewidth]{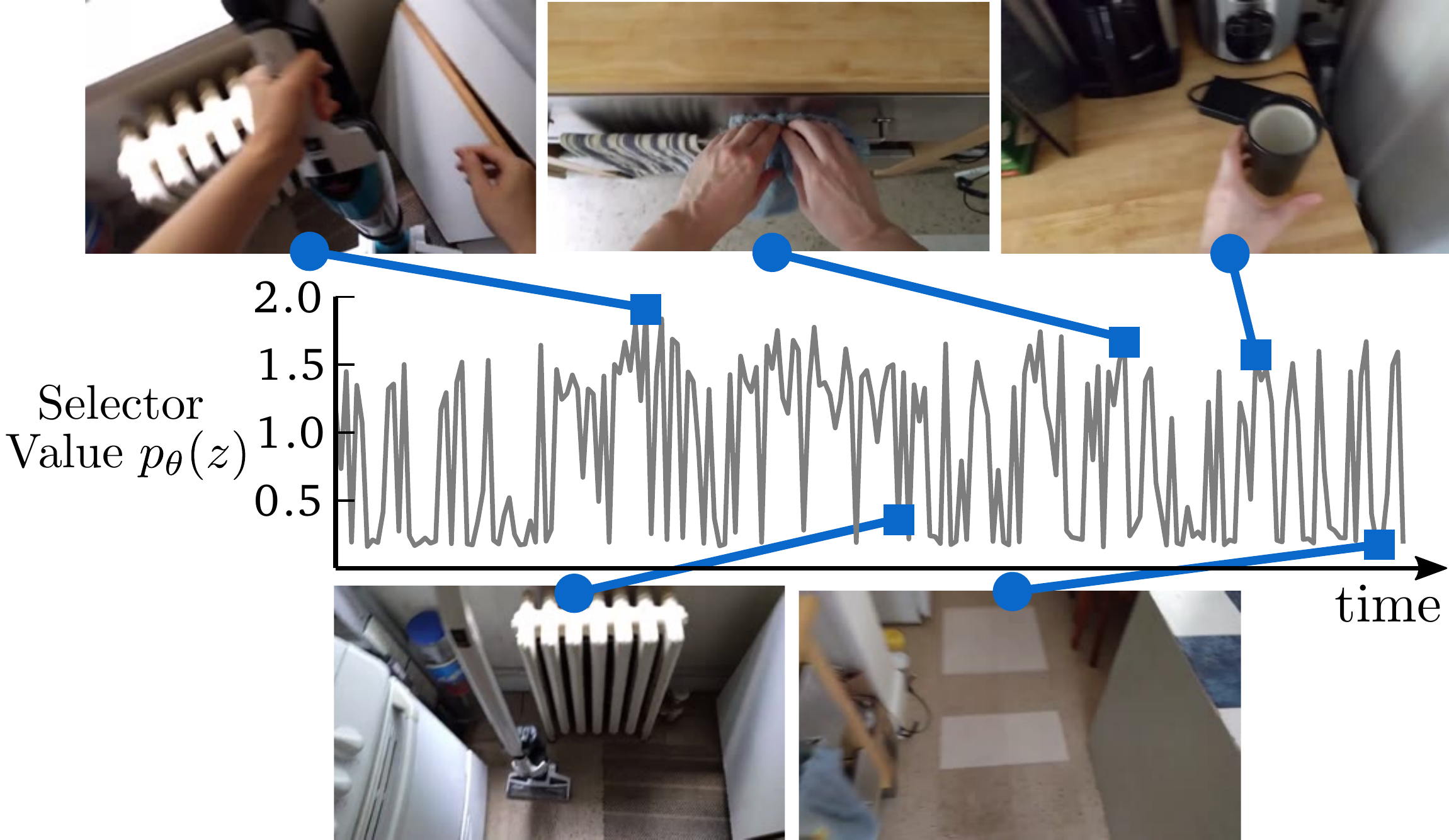}
    \vspace{-0.7cm}
    \caption{Our model learns to assign weights to all the frames in both third and first-person videos. Here we show the selector value $p_\theta(z)$ (the importance of each frame) for a sample first-person video, and highlight frames assigned with high and low values. See Section~\ref{sec:map3to1} for details.\vspace{-0.2cm}}
    \label{fig:videovisual}
\end{figure}

Quantitatively, we found that $68\%$ of high-ranked and only $15\%$ of low-ranked frames contained hands. This is further highlighted in Figures~\ref{fig:attention} and~\ref{fig:attentionpairs} where we visualize conv5 activations, and gradients at the image layer, respectively. We observe the network attending to hands, objects, and the field of view. Figure~\ref{fig:videovisual} illustrates the selection over a video sequence. Here, we include the selector value of $p_\theta(z)$ for each frame in a first-person video. The images highlight points in the graph with particularly useful/useless frames. In general, we see that the weights vary across the video, but the high points correspond to useful moments in the first-person video (top row of images), for example, with a clear view of hands manipulating objects.

\subsection{Alignment and localization}
\label{sec:align}
In the second experiment we align a given first-person moment in time, i.e., a set of frames in a one-second time interval, with a third-person video, and evaluate this temporal localization. In other words, our task is to find any one-second moment that is shared between those first and third-person perspectives, thus capturing their \emph{semantic similarity}. This allows for evaluation despite uninformative frames and approximate alignment. For evaluation, we assume that the ground truth alignment can be approximated by temporally scaling the first-person video to have the same length as the third-person video.

If $m$ denotes all the possible one-second moments in a first-person and $n$ in a third-person video, there are $m\times n$ ways to pick a pair of potentially aligned moments. Our goal is to pick the pair that has the best alignment from this set. The moments are shuffled so there is no temporal context. We evaluate this chosen pair by measuring how close these moments are temporally, in seconds, as shown in Table~\ref{tbl:alignment}. To this end, we use our learned model, and find  one-second intervals in both videos that have the lowest sum of distances between the frames within this moment. We use L2 distance between fc7 features in these experiments.

We present our alignment results in Table~\ref{tbl:alignment}, and compare with other methods. These results are reported as median alignment error in seconds. The performance of fc7 features from the ImageNet ResNet-152 network is close to that of a random metric ($11.0$s). `Two-Stream', which refers to the performance of RGB features from the two-stream network trained on the Charades dataset, performs better. Our `ActorObservetNet' outperforms all these methods.

\begin{figure}[tb]
    \centering
    \includegraphics[width=1.0\linewidth]{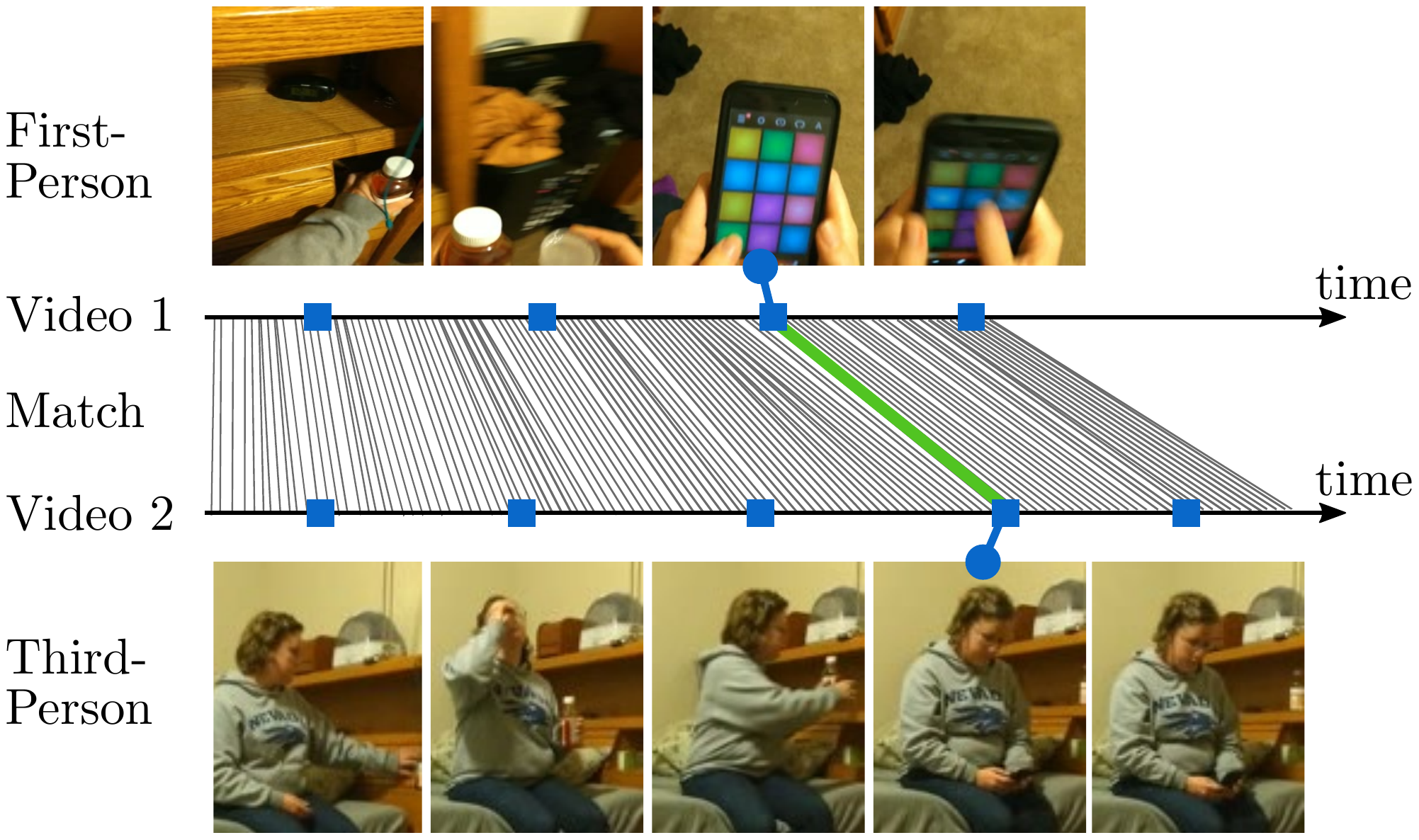}
    \vspace{-0.6cm}
    \caption{Our model matches corresponding moments between two videos. We find the moment in the third-person video (bottom row) that best matches (shown in green) our one second first-person moment (top row), along with other possible matches (gray). ({\it Best viewed in pdf.})\vspace{-0.2cm}}
    \label{fig:alignment}
\end{figure}

\new{
\begin{table}[tb]
\begin{center}
\resizebox{\linewidth}{!}{%
\begin{tabular}{@{}lccccc@{}}
\toprule
 & Random Chance & Human & \begin{tabular}[c]{@{}l@{}}ImageNet\\ ResNet-152\end{tabular} & \begin{tabular}[c]{@{}l@{}}Charades\\ Two-Stream\end{tabular} & ActorObserverNet \\ \midrule
Same person & 11.0 & 1.3 & 8.3 & 6.5 & \bf{5.2} \\ 
Different persons & 11.0 & 1.3 & 8.7 & 7.0 & \bf{6.1} \\ \bottomrule
\end{tabular}%
}
\end{center}
\vspace{-0.4cm}
\caption{Alignment error in seconds for our method `ActorObserverNet' and baselines. Lower is better. See Section~\ref{sec:align} for details.\vspace{-0.3cm}}
\label{tbl:alignment}
\end{table}
}
We visualize the temporal alignment between a pair of videos in Figure~\ref{fig:alignment}. We highlight in green the best moment in the video chosen by the model: the person looking at their cell phone in the third-person view, and a close-up of the cell phone in the first-person view.

\subsection{Zero-shot first-person action recognition}
\label{sec:zeroshot}
Since our ActorObserverNet model learns to map between third and first-person videos, we use it to transfer knowledge acquired from a dataset of third-person videos, annotated with action labels, to the first-person perspective. In essence, we evaluate first-person action recognition in a zero-shot setting. We annotated first-person videos in the test set with the 157 categories from Charades~\cite{charades} to evaluate this setup. Following the evaluation setup from Charades, we use the video-level multi-class mean average precision (mAP) measure.

\begin{table}[tb]
\begin{center}
\resizebox{\linewidth}{!}{%
\begin{tabular}{@{}lcccc@{}}
\toprule
 & Random & \begin{tabular}[c]{@{}l@{}}Charades\\ VGG-16\end{tabular} & \begin{tabular}[c]{@{}l@{}}Charades\\ ResNet-152\end{tabular} & ActorObserverNet \\ \midrule
Accuracy & 8.9 & 17.8 & 22.7 & \bf{25.9} \\ \bottomrule
\end{tabular}
}
\end{center}
\vspace{-0.5cm}
\caption{Egocentric action recognition in the zero-shot learning setup. We show the video-level mAP on our Charades-Ego dataset. Higher is better. See Section~\ref{sec:zeroshot} for details.\vspace{-0.3cm}}
\label{tbl:zerofirst}
\end{table}
In order to transfer knowledge from the third-person to the first-person perspective, we add a classification loss to the third-person model after the fc7 layer. To train this framework, we use third-person training examples from the Charades dataset, in addition to the training set from our Charades-Ego dataset. Note that the third-person videos from Charades are annotated with action labels, while our data only has unlabelled first/third person pairs. Thus, we use the mapping loss in (\ref{eq:finalloss}) when updating the network parameters due to first/third person pair, and the RGB component of the two-stream classification loss for an update due to a Charades third-person example.

Our model now learns to not only map both first and third-person frames to a shared representation, but also a third-person activity classifier on top of that shared representation. At test time, we make a prediction for each frame in a first-person test video, and then combine predictions over all the video frames with mean pooling. We present the results in Table~\ref{tbl:zerofirst}.

\vspace{0.2cm}
\noindent{\bf Baseline results.} The performance of random chance is $8.9\%$ on the Charades-Ego dataset. We also compare to the RGB two-stream model trained on Charades (third-person videos), using both VGG-16 and ResNet-152 architectures, which achieve $18.6\%$ and $22.8\%$ mAP respectively, on the Charades test set. Both are publicly available~\cite{charades}, and show a $8.9\%$ and $13.8\%$ improvement respectively, over random chance on our first-person videos.

\vspace{0.2cm}
\noindent{\bf Our results.} Our ActorObserverNet further improves over the state-of-the-art two-stream network by $3.2\%$. This shows that our model can transfer knowledge effectively from the third-person to the first-person domain. 

To further analyze whether the gain in performance is due to a better network, or third to first-person transfer, we evaluated our network on the Charades test set. It achieves $23.5\%$ on third-person videos, which is only $0.7\%$ higher than the original model, which suggests that the performance gain is mainly due to the new understanding of how third-person relates to first-person view.

\section{Summary}
We proposed a framework towards linking the first and third-person worlds, through our novel Charades-Ego dataset, containing pairs of first and third-person videos. This type of data is a first big step in bringing the fields of third-person and first-person activity recognition together. Our model learns how to jointly represent those two domains by learning a robust triplet loss. Semantic equivalence in data allows it to relate the two perspectives from different people. Our results on mapping third-person to first-person, alignment of videos from the two domains, and zero-shot first-person action recognition clearly demonstrate the benefits of linking the two perspectives.

\paragraph{Acknowledgments.} This work was supported by Intel via the Intel Science and Technology Center for Visual Cloud Systems, Sloan Fellowship to AG, the Inria associate team GAYA, the ERC advanced grant ALLEGRO, gifts from Amazon and Intel, and the Indo-French project EVEREST (no.\ 5302-1) funded by CEFIPRA. The authors would like to thank Achal Dave, Vicky Kalogeiton, Kris Kitani, Nick Rhinehart, {Jardin du Th\'e} for their invaluable suggestions and advice, and the Amazon Mechanical Turk workers for their time.

{\small
\bibliographystyle{ieee}
\bibliography{CVPR2018_Gunnar}
}
\newpage
\onecolumn



\section{Supplementary Material}

\noindent This supplementary material contains the following.

\begin{enumerate}
\item Details of the implementations of the new layers 
\item Details of ActorObserverNet
\item Full derivation of the loss with respect to the selector
\end{enumerate}


\subsection{Implementation of the new layers}

In this section we derive the equations that are used to update the VideoSoftmax layer (Eq.~3 from the paper) and final loss layer (Eq.~2 and Eq.~4 from the paper) with SGD. This is needed since the equations require computations across samples in different batches (frames in the same video across batches) and we need to do the computation in an online fashion.

\paragraph{VideoSoftmax}

We want to implement the VideoSoftmax layer to fit the SGD framework. We start with the VideoSoftmax objective (Eq.~3 from the paper):
\begin{align}
p_\theta(x) &= \frac{e^{f_\theta(x)}}{\sum\limits_{x \in \mathcal{V}} e^{f_\theta(x)}}, \\ 
&= \frac{e^{f_\theta(x)}}{e^{f_\theta(x)} + \sum\limits_{\tilde{x} \in \mathcal{V} \smallsetminus x} e^{f_\theta(\tilde{x})}}.
\end{align}
We now make the normalization explicit:
\begin{align}
p_\theta(x) &= 
\frac{1}{N} \frac{e^{f_\theta(x)}}{\frac{1}{N}e^{f_\theta(x)} + \frac{N-1}{N} \frac{1}{N-1} \sum\limits_{\tilde{x} \in \mathcal{V} \smallsetminus x} e^{f_\theta(\tilde{x})}}, \\ 
p_\theta(x) &= 
k \frac{e^{f_\theta(x)}}{k e^{f_\theta(x)} + (1-k) \Sigma_{N-1}^\mathcal{V}},
\end{align}
where $N$ is the number of terms in the sum. We replace this with a constant $k$ that is defined to be $k=0.1$. Here $\Sigma_N^\mathcal{V}{=}\frac{1}{N}\sum\limits_{\tilde{x} \in \mathcal{V} \smallsetminus x} e^{f_\theta(\tilde{x})}$. To avoid having $p_\theta$ of different ranges for different hyperparameters, we work with $p_\theta(x)/k$ which has the expected value of $1$. Our final online update equation is as follows:
\begin{align}
    \Sigma_N^\mathcal{V} &= k e^{f_\theta(x)} + (1{-}k) \Sigma_{N-1}^\mathcal{V}, \\ 
    \frac{p_\theta(x)}{k} &= \frac{e^{f_\theta(x)}}{\Sigma_N^\mathcal{V}},
\end{align}
where $\Sigma_N^\mathcal{V}$ is our online update of the denominator (the normalization constant) for video $\mathcal{V}$.

\paragraph{Loss layer}
\def\xzz{\tau}
\def\hxzz{\hat{\tau}}
\def\txzz{\tilde{\tau}}\def\df{\frac{\partial }{\partial f(\hxzz)}}
\def\sumxh{\sum_{\xzz \smallsetminus \hxzz}}
\def\sumh{\sum\limits_{\hxzz}}
\def\sumt{\sum\limits_{\txzz}}
\def\efx{e^{f(\xzz)}}
\def\efh{e^{f(\hxzz)}}
\def\eft{e^{f(\txzz)}}

The details of the online equation for the loss are slightly more involved than the previous VideoSoftmax because of the importance sampling, but results in a similarly simple equation. For clarity we use a shorthand notation for the triplet $\xzz{=}(x{,}z{,}z')$.
We start with the loss from Eq.~2 from the paper and write out the normalization constant explicitly:
\def\hatxxz{\hxzz}
\begin{align}
L &= \sum_{\xzz \sim Q} \frac{p_{\theta}(\xzz)}{q(\xzz)} l_{\theta}(\xzz), \\ 
&= N \frac{|Q|}{N} \sum_{\xzz \sim Q} p_{\theta}(\xzz) l_{\theta}(\xzz), \\
&\approx N \frac{1}{\sum_{\xzz \sim Q} p_{\theta}(\xzz)} \sum\limits_{\xzz \sim Q} p_{\theta}(\xzz) l_{\theta}(\xzz) ,
\end{align}
where $\xzz {\sim} Q$ indicates that $\xzz$ is sampled from $Q$, $N$ is the number of samples we draw from $Q$, and $|Q|$ is the size of $Q$. We can write $q(\xzz)=\frac{1}{|Q|}$ because $Q$ is uniform. Finally, we use $\frac{1}{|Q|}{\approx}\frac{\sum_{\xzz \sim Q} p_{\theta}(\xzz)}{N}$ (normalized importance sampling).
We then break the sum into parts for the current sample $\hxzz$ and other samples $\xzz$ and work with $\frac{L}{N}$:
\def\sumqp{\sum\limits_{\xzz \sim Q} p_{\theta}(\xzz)}
\def\sumqpl{\sumqp l_{\theta}(\xzz)}
\def\pl{p_{\theta}(\xzz) l_{\theta}(\xzz)}
\def\hsumqp{\sum\limits_{\txzz \sim Q} p_{\theta}(\txzz)}
\def\hsumqpl{\hsumqp l_{\theta}(\txzz)}
\def\sumpqp{p_{\theta}(\xzz) + \hsumqp}
\def\ksumpqp{kp_{\theta}(\xzz) + (1{-}k)\hsumqp}
\def\sksumpqp{kp_{\theta}(\xzz) + (1{-}k)\Sigma_{N-1}}
\begin{align}
\frac{L}{N} &= 
\frac{1}{\sumqp} \left( \pl + \hsumqpl \right), \\
&= 
\frac{1}{\sumpqp} \left( \pl + \hsumqpl \right), \\
&= 
\frac{1}{\frac{\sumpqp}{N}} \left( \frac{1}{N} \pl + 
\frac{N{-}1}{N}\frac{1}{N{-}1} \hsumqpl \right),  \\
&= 
\frac{1}{\sksumpqp} \left( k\pl + 
(1{-}k) \frac{\hsumqpl}{(N{-}1)\hsumqp} \right),
\end{align}
where we scaled the by $\frac{N}{N}$ and $\frac{N-1}{N-1}$ to write the recursive equation. We can see that the right hand side contains $\frac{L}{N-1}$ except with one less samples. We defined $k{=}\frac{1}{N}$ and fix $k{=}0.1$. We define $\Sigma_N=\frac{\hsumqp}{N}$. 
Now we can write this as:
\begin{align}
L_N &= 
\frac{k\pl + (1{-}k)\Sigma_{N-1} L_{N-1}}{\Sigma_N}, \\
\Sigma_N &= \sksumpqp,
\end{align}
where $L_N$ is $\frac{L}{N}$ with $N$ samples drawn from $Q$. This is used to estimate $L$ in order to compute the update in Eq.~4 from the paper.

\subsection{Details of the final triplet network}

In this section we describe additional implementation details of the model. The third-person classification loss attaches to the third-person stream when it is used. When in use, the losses are toggled on or off depending if they have training data in the given triplet. In the classification mixed setup the triplet contains either $(x,\emptyset,\emptyset)$ along with a classification label, otherwise we have $(x,z,z')$ and no classification label.
The batchsize was set to 15 to accomodate the 3 ResNet-152 streams. We used a $3e^{-5}$ learning rate and reduced the learning rate by $10$ every $3$ epochs. We used momentum of $0.95$. 
Since our implementation has two very different loss updates (triplet loss, selector update, and classification loss) we found it initially difficult balance the losses. This was solved by rescaling all gradients to have norm equal to the triplet loss gradient.

To balance the ability of the model to choose what samples to learn from and overfitting we introduced a Tanh layer to bound the possible $f(x)$ (Eq.~3 from the paper) values in the network. We allowed the model to learn this weight for each $p$ starting from Gaussian noise of $\sigma{=}5$. While sharing of the FC layers and scaling parameters did not affect the results in Section 4.2, we found the network to have better performance in Section 4.3 and Section 4.4 when sharing parameters between the two first-person FC streams, and constraining the TanH scale to be positive. This is likely because it adds additional constraints on the selector, and discourages it from overfitting.

To ensure consistent training and testing setup. The triplets from $Q$ consist of every third-person frame, paired with the best corresponding frame (alignment error is estimated to be approximately a second, which implies $\Delta{=}1$ sec), as well as a randomly sample noncorresponding frame that it at least 10 seconds away.

\subsection{Full derivation of the loss with respect to the selector}
The loss in Eq.~2 includes contribution from $p_\theta$, where each output of $p_\theta$ (VideoSoftmax) is normalized across all frames in that video.  
We assume that the last layer before the loss layer is a softmax layer (i.e. VideoSoftmax):
\begin{align}
    p_\theta(\xzz) &= \frac{e^{f(\xzz)}}{\sum\limits_{\txzz} e^{f(\txzz)}}, \\
    &= \frac{e^{f(\xzz)}}{e^{f(\xzz)} + \sum\limits_{\txzz} e^{f(\txzz)}}.
\end{align}
This allows us to account for the contribution of $e^{f(\xzz)}$ to other samples. That is, note that the denominator includes the values over other frames (in the same video), so those terms have to be included in the derivative, the second equation clarifies this relationship by separating the triplet of interest from the sum. For clarity we again use a shorthand notation for the triplet $\xzz{=}(x{,}z{,}z')$.

We start with Eq.~2 from the paper, and insert the assumption for $p_\theta$:
\begin{align}
    L &= \sum_{\xzz} p_{\theta}(\xzz) l_{\theta}(\xzz), \\ 
    &= \sum_{\xzz} \frac{\efx}{\sumt \eft} l_{\theta}(\xzz),
\end{align}
where we use $\txzz$ to emphasize different triplets. We take the derivative of this with respect to $f(\hxzz)$, a particular input to the Softmax that occurs once in the numerator and many times in the denominator:
\begin{align}
    \frac{\partial L}{\partial f(\hxzz)} 
    &= \df \sum_{\xzz} \frac{\efx}{\sumt \eft} l_{\theta}(\xzz), \\
    &= \df \frac{\efh}{\sumt \eft} l_{\theta}(\hxzz) 
       + \sumxh \frac{\efx}{\sumt \eft} l_{\theta}(\xzz), \\
    &= \df \frac{\efh}{\efh + \sumxh \efx} l_{\theta}(\hxzz) 
       + \df \sumxh \frac{\efx}{\efh+\sumxh \efx} l_{\theta}(\xzz),
\end{align}
where we have expanded the softmax to make clear where $\hxzz$ occurs in the numerator and denominator. That is, $\xzz \smallsetminus \hxzz$ is the set of all $\xzz$ that excludes $\hxzz$. We then find the derivative similarly to the derivative of softmax, where we write it in terms of $p_\theta(\hxzz)$ for clarity:
\begin{align}
    \frac{\partial L}{\partial f(\hxzz)} 
    &= p_\theta(\hxzz)(1{-}p_\theta(\hxzz))l_\theta(\hxzz) + \sumxh p_{\theta}(\hxzz)(-p_\theta(\xzz))l_\theta(\xzz), \\
    &= p_\theta(\hxzz)l_\theta(\hxzz) - p_\theta(\hxzz)p_\theta(\hxzz)l_\theta(\hxzz) + \sumxh p_{\theta}(\hxzz)(-p_\theta(\xzz))l_\theta(\xzz), \\
    &= p_\theta(\hxzz)l_\theta(\hxzz) - p_\theta(\hxzz) \left( p_\theta(\hxzz)l_\theta(\hxzz) + \sumxh p_{\theta}(\hxzz)l_\theta(\xzz) \right), \\
    &= p_\theta(\hxzz)l_\theta(\hxzz) - p_\theta(\hxzz) L, \\
    &= p_\theta(\hxzz)(l_\theta(\hxzz) - L).
\end{align}
Here we factorize and recombine the terms that constitute the definition of $L$, allowing us to write this in a compact format. These terms are then implemented in an online fashion as previously described. The final update is Eq.~4 in the paper:
\begin{align}
    \frac{\partial L}{\partial f(x{,}z{,}z')} &= p_{\theta}(x{,}z{,}z')(l_{\theta}(x{,}z{,}z')-L).
\end{align}


\end{document}